\pdfoutput=1

\documentclass[11pt]{article}

\usepackage{emnlp2021}

\usepackage{times}
\usepackage{latexsym}
\usepackage{multirow}
\usepackage{graphicx}
\usepackage{algorithm}
\usepackage{algorithmic}
\usepackage{amsmath}
\usepackage{url}
\usepackage{comment}
\usepackage{enumitem}
\usepackage{color}
\usepackage{subfigure}
\usepackage[T1]{fontenc}
\usepackage{footnote}
\usepackage{hyperref}
\usepackage{listings}
\usepackage{booktabs}

\usepackage{xcolor}
\usepackage{threeparttable}

\definecolor{codegreen}{rgb}{0,0.6,0}
\definecolor{codegray}{rgb}{0.5,0.5,0.5}
\definecolor{codepurple}{rgb}{0.58,0,0.82}
\definecolor{backcolour}{rgb}{0.95,0.95,0.92}

\lstdefinestyle{mystyle}{
    backgroundcolor=\color{backcolour},   
    commentstyle=\color{codegreen},
    stringstyle=\color{codepurple},
    basicstyle=\ttfamily\scriptsize,
    breakatwhitespace=true,         
    breaklines=true,                 
    captionpos=b,                    
    keepspaces=true,                 
    numbers=none,                    
    numbersep=5pt,                  
    showspaces=false,                
    showstringspaces=false,
    showtabs=false,                  
    tabsize=2,
    columns=flexible,
    escapeinside={(*}{*)},
}

\usepackage{amssymb}
\usepackage[utf8]{inputenc}

\usepackage{microtype}
\usepackage[hang,flushmargin]{footmisc}

\usepackage[misc]{ifsym}

\title{EasyPhoto: Your Personal AI Photo Generator}

\author{Ziheng Wu$^{1}$\thanks{* Equal Contribution. $^{(\textrm{\Letter})}$ Corresponding Author.}, Jiaqi Xu$^{1*}$, Xinyi Zou$^{1}$, Kunzhe Huang$^{1}$, Xing Shi$^{1}$$^{(\textrm{\Letter})}$, Jun Huang$^{1}$$^{(\textrm{\Letter})}$\\
$^1$ Platform of AI (PAI), Alibaba Group\\
\texttt{\{zhoulou.wzh, zhoumo.xjq, zouxinyi.zxy, huangkunzhe.hkz,}\\
\texttt{shubao.sx, huangjun.hj\}@alibaba-inc.com}
}

\begin{document}
\maketitle

\begin{abstract}
\href{https://github.com/AUTOMATIC1111/stable-diffusion-webui}{Stable Diffusion web UI (SD-WebUI)}  is a comprehensive project that provides a browser interface based on Gradio library for Stable Diffusion models. 
In this paper, We propose a novel WebUI plugin called EasyPhoto, which enables the generation of AI portraits. By training a digital doppelganger of a specific user ID using 5 to 20 relevant images, the finetuned model (according to the trained LoRA model) allows for the generation of AI photos using arbitrary templates. Our current implementation supports the modification of multiple persons and different photo styles. Furthermore, we allow users to generate fantastic template image with the strong SDXL model, enhancing EasyPhoto's capabilities to deliver more diverse and satisfactory results. The source code for EasyPhoto is available at: \url{https://github.com/aigc-apps/sd-webui-EasyPhoto}\footnote{We also support a webui-free version by using  diffusers: https://github.com/aigc-apps/EasyPhoto}. We are continuously enhancing our efforts to expand the EasyPhoto pipeline, making it suitable for any identification (not limited to just the face), and we enthusiastically welcome any intriguing ideas or suggestions.

\end{abstract}

\section{Introduction}

\begin{figure*}[h]
	\centering
	\includegraphics[width=.95 \linewidth]{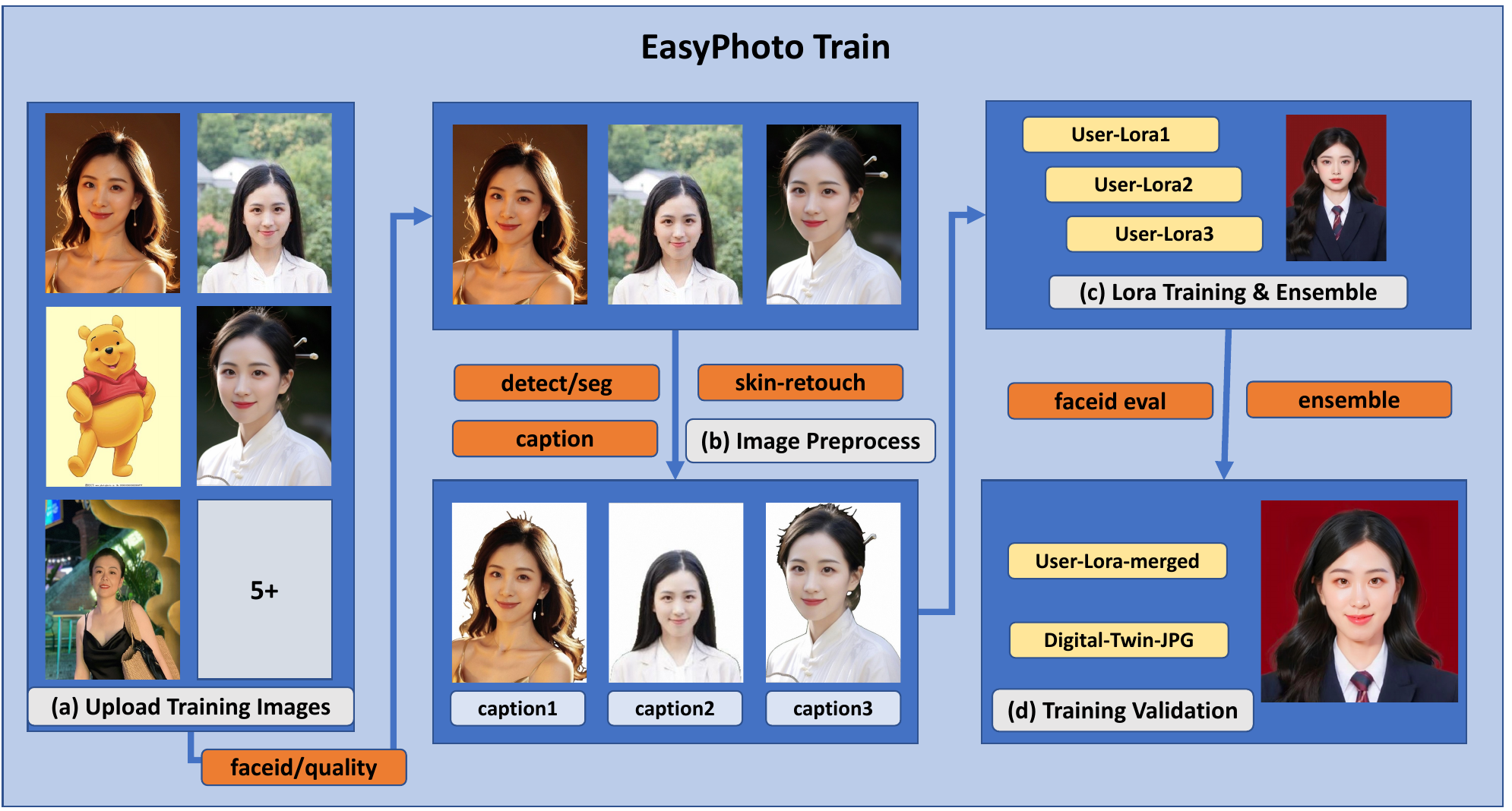}
	\caption{The pipeline of the training process in EasyPhoto, including: (a) Upload Training Images, (b) Image Preprocess, (c) LoRA models Training \& Ensemble, (d) Training Validation.}\label{fig:train}
\end{figure*}

\begin{figure*}[h]
	\centering
	\includegraphics[width=1.0 \linewidth]{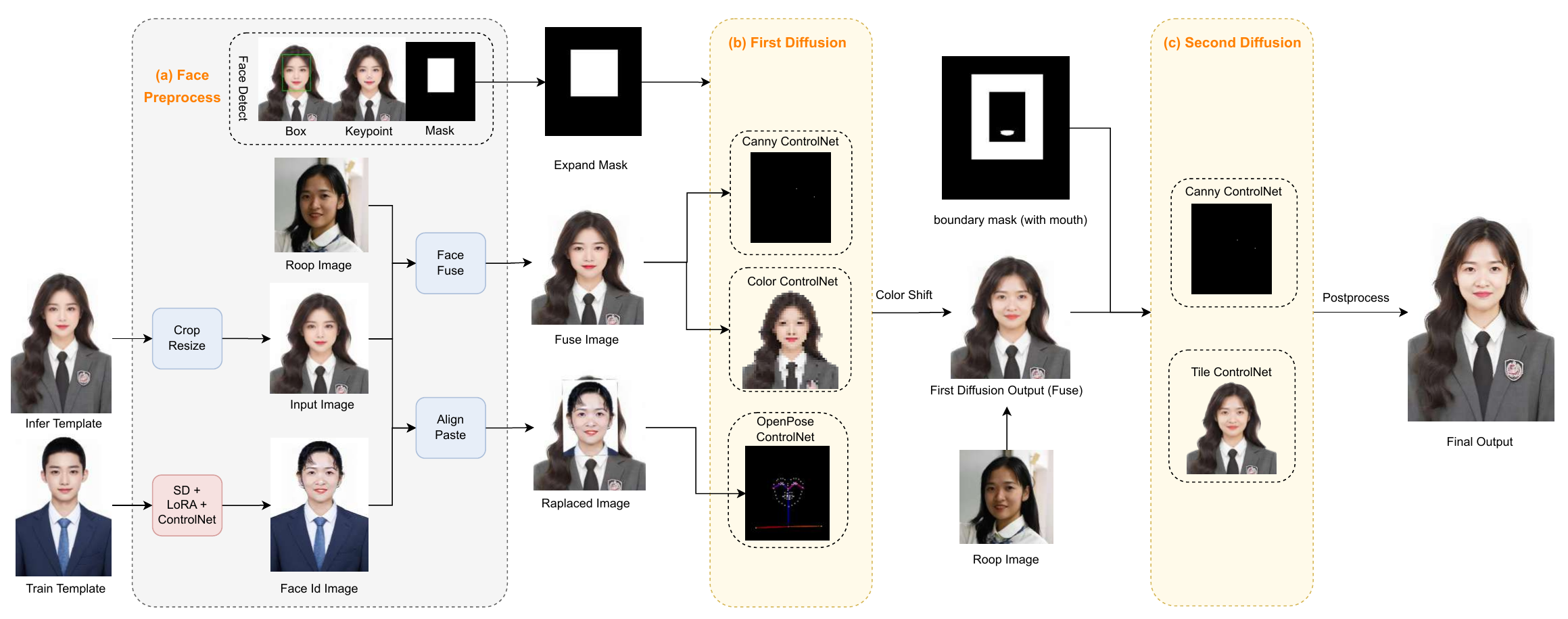}
	\caption{The pipeline of the inference process in EasyPhoto, including: (a) Face Preprocess to obtain the preprocessed input image, (b) First Diffusion to generate image resemble to the input user, (c) Second Diffusion to fix boundary artifacts problem.}\label{fig:pipe}
\end{figure*}

Stable Diffusion \cite{rombach2021highresolution} is indeed a popular diffusion-based generation model that is widely used for generating realistic images based on text descriptions. It has applications in various domains such as image-to-image translation, image inpainting, image outpainting, \emph{etc}.

Unlike other models like DALL-E\footnote{https://openai.com/dall-e-2} and Midjourney\footnote{https://www.midjourney.com/}, Stable Diffusion is open source, making it highly flexible for further development. One of the most well-known applications of Stable Diffusion is the Stable Diffusion web UI. It includes a browser interface, built on the Gradio library, providing a user-friendly interface for Stable Diffusion models (SD models). It integrates various SD applications, preprocess functions to enhance the usability and control of image generation. 

Due to the convenience of SD-WebUI, we develop EasyPhoto as a WebUI plugin to generate AI portraits. Unlike existing approaches that may introduce unrealistic lighting or suffer from identity loss, EasyPhoto uses the image-to-image capability of the SD model to ensure realistic and accurate results. EasyPhoto can be easily installed as an extension in WebUI, making it user-friendly and accessible to a wide range of users. By leveraging the power of the SD model, EasyPhoto enables users to generate high-quality, identity-guided AI portraits that closely resemble the input identity.

First, we allow user to upload several images to online train a face LoRA model as the user's digital doppelganger. The LoRA (Low-Rank Adaptation \cite{hu2022lora}) model is trained using low-rank adaptation technology to quickly fine-tune diffusion models that enable the base model to understand the specific user ID information. These trained models are then integrated and merge into the base SD model for inference. 

During the inference stage, we intend to repaint the facial region in the inference template by stable diffusion models. Several ControlNet \cite{zhang2023adding} units are used to verify the similarity between the input and output image. To overcome issues such as identity loss and boundary artifacts, a two-stage diffusion process is employed. This ensures that the generated images maintain the user's identity while minimizing visual inconsistencies. Furthermore, we believe that the inference pipeline is not limited to just faces but can also be extended to anything related to user ID. This implies that once we train a LoRA model of an id, we can generate various AI photos. It can be used in various applications, such as virtual try-on. We are continuously working towards achieving more realistic results.

\begin{itemize}
	\item We propose a novel approach to train the LoRA model in EasyPhoto by intergrating mulitple LoRA models to maintain the face fidelity of the generated results.

        \item We incorporate reinforcement learning methods to optimize LoRA models for facial identity rewards ,which further improve the identity similarity between the generated results and the training images. 
        
	\item We propose a two-stage inpaint-based diffusion process in EasyPhoto, which aims to generate AI photos with high resemblance and aesthetics. The image preprocessing is thoughtfully designed to create suitable guidance for ControlNet. EasyPhoto enables users to acquire personalized AI portraits with various styles or multiple individuals. Additionally, we leverage the powerful SDXL model to generate templates, resulting in a wide range of diverse and realistic output.
	
	
\end{itemize}

\section{Training Process}

The pipeline of the training process for EasyPhoto can be seen in Fig.\ref{fig:train}.

First, we perform face detection \cite{serengil2020lightface,serengil2021lightface} on the input user image to identify the location of the face. Once the face is detected, the input image is cropped based on a specific ratio to focus solely on the face region. Then, the saliency detection model\footnote{https://modelscope.cn/models/damo/cv\_u2net\_salient-detection/summary}\cite{Qin_2020_PR} and the skin beautification model \footnote{https://modelscope.cn/models/damo/\\cv\_unet\_skin\_retouching\_torch/summary}\cite{lei2022abpn} are utilized to obtain a clean face training image. These models help enhance the visual quality of the face and ensure that the resulting training image predominantly contains the face (i.e., the background information has been removed). The input prompt is fixed as ``easyphoto\_face, easyphoto, 1person''. Experiments show that even with the fixed input prompt, the trained LoRA model can be satisfactory.  Finally, the LoRA model is trained using these processed images and the input prompt, equipping it to comprehend user-specific facial characteristics effectively.

During the training process, we incorporate a critical validation step, where we compute the face id gap between the verification image (image generate by the trained LoRA model and a canny ControlNet based on training template) and the user's image. This validation procedure is instrumental in achieving the fusion of LoRA models, ultimately ensuring that the trained LoRA model becomes a highly accurate digital representation, or doppelganger, of the user. In addition, the verification image with the best face\_id score will be chosen as the face\_id image, which will be used to enhance the identity similarity of the inference generation.

Based on the model ensemble process, the LoRA models are trained using maximum likelihood estimation as the objective, whereas the downstream objective is to preserve facial identity similarity. To bridge this gap, we harness reinforcement learning techniques to directly optimize the downstream objective. Specifically, we define the reward model as the facial identity similarity between the training images and the generated results by LoRA models. We employ DDPO \cite{black2023training} to fine-tune the LoRA models in order to maximize this reward. Consequently, the facial features learned by the LoRA models exhibit improvements, leading to enhanced similarity between template-generated results and demonstrating their generalizability across different templates.


\section{Inference Process}

The inference process of a single User Id in EasyPhoto can be seen in Fig.\ref{fig:pipe}. It consists of three parts, (a) Face Preprocess to obtain the preprocessed input image and ControlNet reference, (b) First Diffusion to generate the coarse result that is resemble to the user input, (c) Second Diffusion to fix the boundary artifacts that make the result more realistic. It takes as input an inference template and a face\_id image, which is generated during training validation with best face\_id score. The result is a highly detailed AI portrait of the input user, closely resembling the user's unique appearance and identity based on the infer template. We will elaborately illustrate each process in this section.

\subsection{Face PreProcess}

An intuitive way to generate an AI portrait based on the inference template is to inpaint the face region of the template with the SD model (LoRA finetuned). Moreover, incorporating ControlNet into this process can significantly enhance the preservation of user identity and similarity within the generated images. However, directly applying the ControlNet for inpainting the region may introduce potential artifacts or issues, which can include:

\begin{itemize}
	\item \textbf{Inconsistency between the template image and the input image.} It is obvious that the key point in the template image are incompatible with those of the face\_id image. Consequently, utilizing the face\_id image as a reference for ControlNet will inevitably lead to inconsistencies in the final result.
	
	\item \textbf{Defects of the inpaint region.} Masking the region and inpainting it with a new face will result in noticeable defects, particularly along the inpaint boundary. This will negatively impact the realism and authenticity of the generated result.
    
	\item \textbf{Identity Loss by ControlNet.} As ControlNet is not utilized in the training process, incorporating ControlNet during the inference process may compromise the ability of the trained LoRA models to preserve the identity of the input user id.
	
\end{itemize}

To address the aforementioned issues, we propose three procedures for face preprocessing in order to obtain the input image, mask and reference image for the first diffusion stage.

\noindent\textbf{Align and Paste.} To address the issue of face landmark mismatch between the template and the face\_id, we propose an affine transformation and face-pasting algorithm. Firstly, we calculate the face landmarks of both the template image and the face\_id image. Next, we determine the required affine transformation matrix $M$ that aligns the face landmarks of the face\_id image with those of the template image. We then directly apply this matrix $M$ to paste the face\_id image onto the template image. The resulting replaced image retains the same landmarks as the face\_id image while aligning with the template image. Consequently, using this replaced image as the reference for the openPose ControlNet ensures the preservation of both the face identity and the face\_id image and the facial structure of the template image.

\noindent\textbf{Face Fuse.} To rectify boundary artifacts caused by mask inpainting, we propose a novel approach involving artifact rectification through the use of a canny ControlNet. This allows us to guide the image generation process and ensure the preservation of harmonious edges. However, compatibility issues may arise between the canny edges of the template face and the target ID. To overcome this challenge, we employ the FaceFusion\footnote{https://modelscope.cn/models/damo/cv\_unet-image-face-fusion\_damo/summary} algorithm to fuse the template and roop image (one of the ground truth user images). By fusing the images, the resulting fused input image exhibits improved stabilization of all edge boundaries, leading to better results during the first diffusion stage. This approach helps to alleviate boundary artifacts and enhance the overall quality of the generated image.

\noindent\textbf{ControlNet-based Validation.} As the LoRA model was trained without ControlNet, using ControlNet during inference may influence LoRA model's identity preservation. To mitigate this issue, we introduce the use of ControlNet for validation in the training pipeline. Validation images are generated by applying the LoRA model and the ControlNet model to standard training templates. We then compare the face\_id score between the validation images and the corresponding training image to integrate the LoRA models effectively. By incorporating models from different stages and considering the influence of ControlNet, we substantially enhance the model's generalization capability. This validation-based approach ensures improved identity preservation and facilitates the seamless integration of ControlNet within the LoRA model during model inference.

\begin{figure*}[th]
	\centering
	\includegraphics[width= \linewidth]{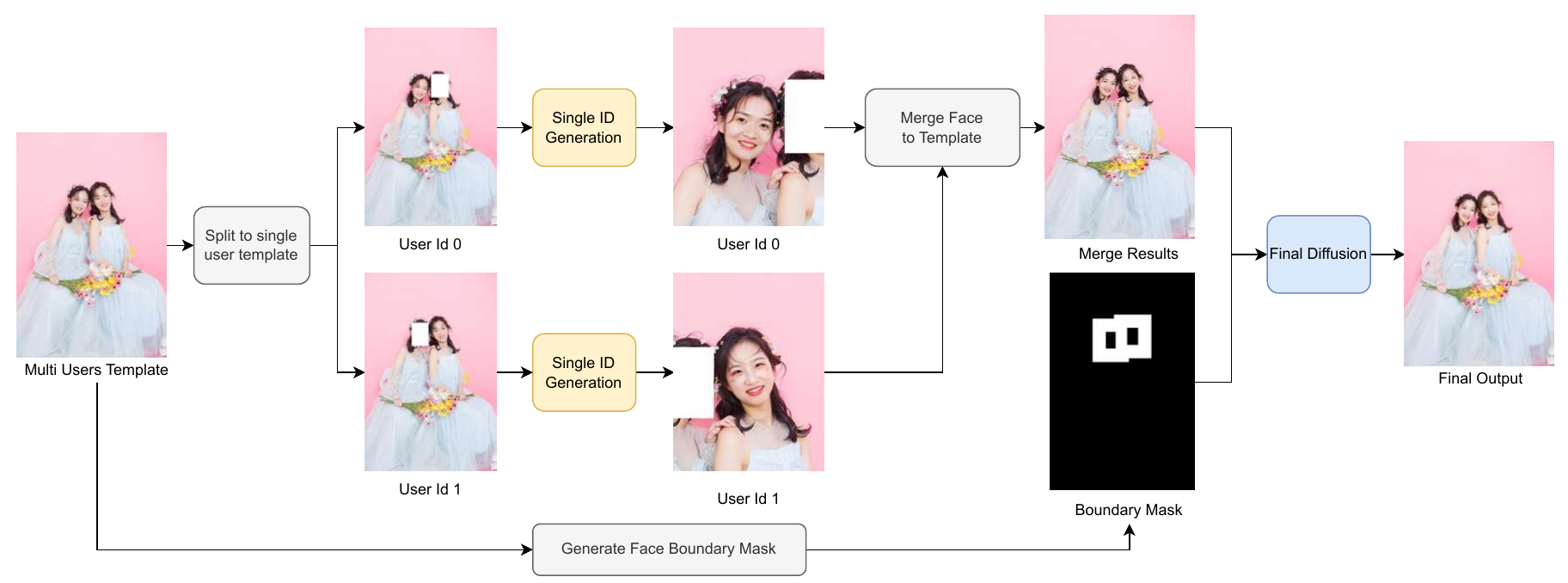}
	\caption{The pipeline of the inference process in EasyPhoto for multi-user IDs. It includes a preprocess mask split process and a final diffusion process to expand the single user ID pipeline to support multiple users.}\label{fig:multi}
\end{figure*}

\subsection{First Diffusion}

The first diffusion stage serves to generate an image with a specific id (similar to the input user id) based on the template image. The input image is a fusion of the template image and the user's image, while the input mask is the calibrated face mask of the template image (expand to include the ears).

To enhance control over the image generation process, three ControlNet units are integrated. The first unit focuses on canny control of the fused image. By utilizing the fused image as a guide for the canny edges, a more precise integration of the template and face\_id image is achieved. Compared to the original template image, the fused image naturally combines the id information in advance, resulting in minimized boundary artifacts. The second ControlNet unit is the color control of the fused image. It verifies the color distribution within the inpainted region, ensuring consistency and coherence. The third ControlNet unit is the openpose control of the replaced image, which contains both the user's face identity and the facial structure of the template image. This guarantees the similarity and stability of the generated images. With the incorporation of these ControlNet units, the first diffusion process can successfully generate a high-fidelity result that closely resembles the user's specified id.

\subsection{Second Diffusion}

The second diffusion stage is dedicated to fine-tuning and refining the artifacts near the face boundary. Additionally, we provide the option to mask the region of the mouth to enhance the generation within that specific area. Similarly to the first diffusion stage, we fuse the result of the user's image (roop image) with the output image from the first diffusion stage to obtain the input image for the second diffusion stage. This fused image also serves as a reference for the canny controlnet, enabling better control over the generation process. Furthermore, we incorporate the tile ControlNet to achieve higher resolution in the final output. It is beneficial to enhance the details and overall quality of the generated image.

In the final step, we perform post-processing on the generated image to adjust it to the same size as the inference template. This ensures consistency and compatibility between the generated image and the template. Additionally, we apply skin retouch algorithms \footnote{https://modelscope.cn/models/damo/\\cv\_unet\_skin\_retouching\_torch/summary}\cite{lei2022abpn} and portrait enhancement algorithms \footnote{https://modelscope.cn/models/damo/cv\_gpen\_image-portrait-enhancement/summary} \cite{yang2021gpen} to further enhance the quality and appearance of the final result. These algorithms help to refine the skin texture and overall visual appeal, resulting in a better and more polished image.

\subsection{Multi User Ids}

\begin{figure}[t]
	\centering
	\includegraphics[width= \linewidth]{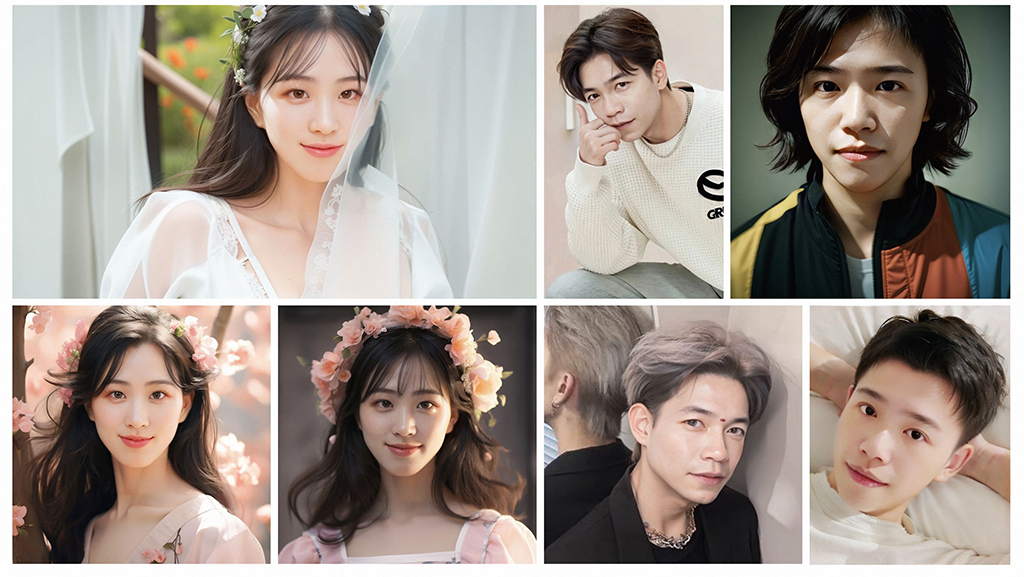}
	\caption{The generation results of EasyPhoto. (Realistic Style base SD model is used)}\label{fig:res1}
\end{figure}

EasyPhoto also supports the generation of multiple user IDs, which is an extension of the single user ID generation process, demonstrated in Fig.\ref{fig:multi}.

To accomplish this, we begin by performing face detection on the inference template. The template is then divided into several masks, with each mask containing only one face while the rest of the image is masked in white. This simplifies the task into individual single user ID generation problems. Once the user id images are generated, they are merged back into the inference template. To address the boundary artifacts that may arise during the merging process, we employ a diffusion-based image inpainting technique with the aid of the face boundary mask. This helps to seamlessly blend the generated images with the template image, resulting in a high-quality group photo.

\begin{figure}[t]
	\centering
	\includegraphics[width=\linewidth]{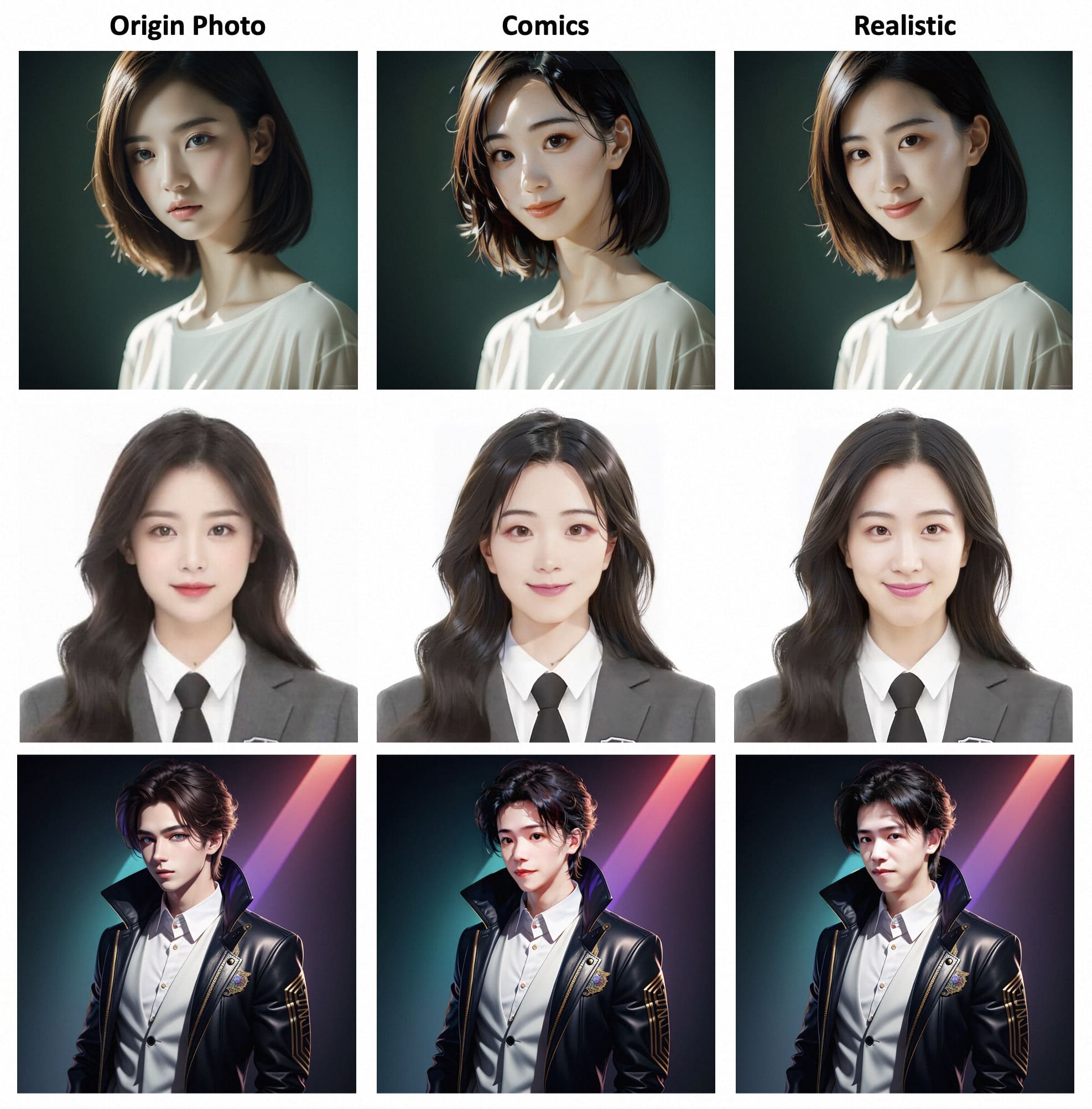}
	\caption{The generation results of EasyPhoto. (Comic Style base SD model is used)}\label{fig:res2}
\end{figure}

\section{Anything Id}

We are currently in the process of expanding the EasyPhoto pipeline to accommodate any ID. This means that users will be able to train a LoRA model for any specific ID (with few training images that include the target object) and utilize the trained model to generate that particular ID. Unlike face-related tasks that have been extensively studied, such as face detection and landmark detection, there is limited existing research focused on general object replacement. The main challenge lies in generating a suitable ControlNet reference for the replacement process. Fortunately, with the availability of powerful general models like SAM \cite{kirillov2023segany}, LightGlue \cite{lindenberger2023lightglue}, and Grounding Dino \cite{liu2023grounding}, it becomes feasible to locate and match the general key objects. We are actively working on updating EasyPhoto to support generation for any ID and will release the code soon.

\section{Experiments}

\begin{figure}[t]
	\centering
	\includegraphics[width=\linewidth]{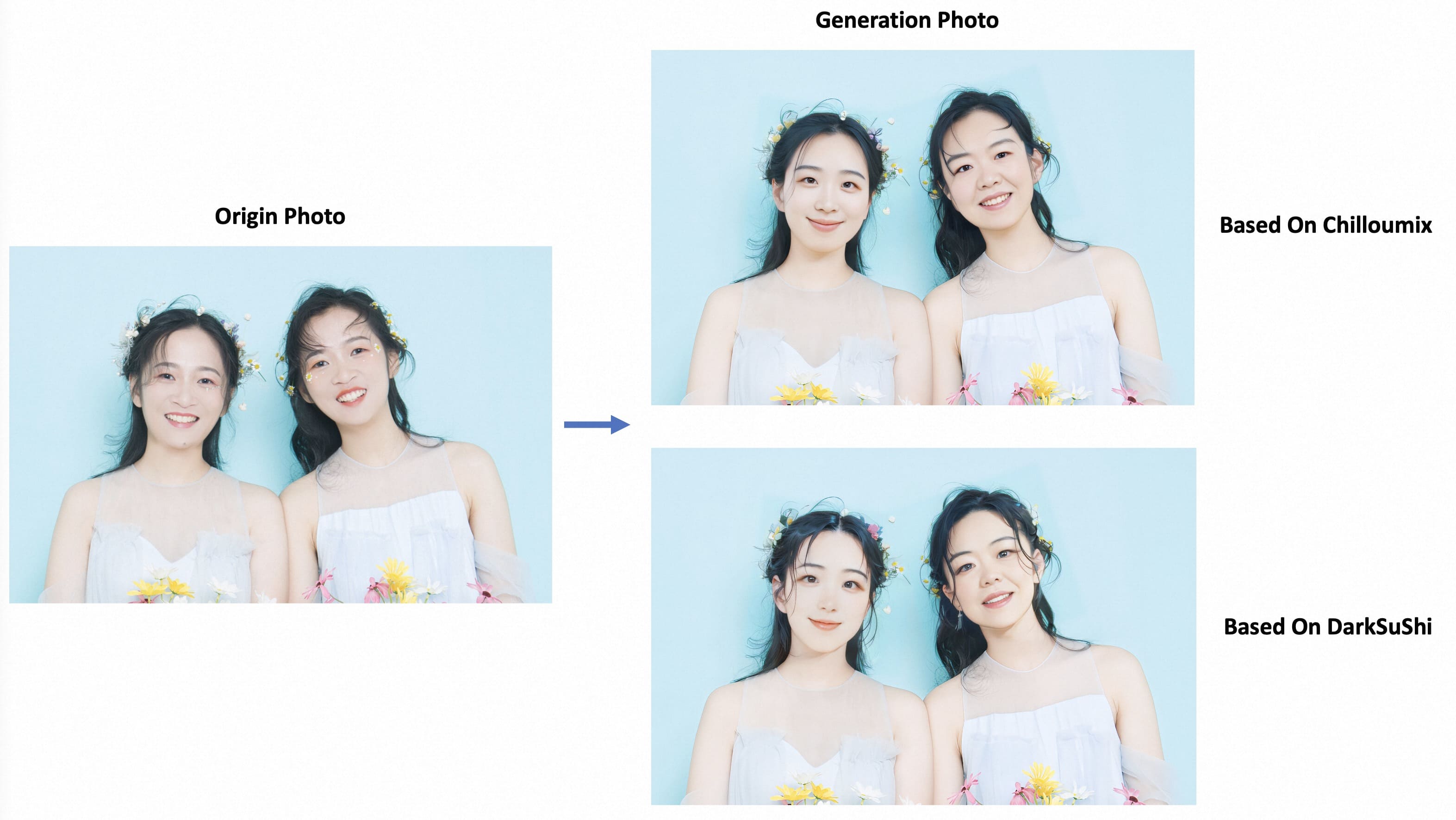}
	\caption{The generation results of EasyPhoto. (Multi Person template is used)}\label{fig:res3}
\end{figure}

\begin{figure}[t]
	\centering
	\includegraphics[width= \linewidth]{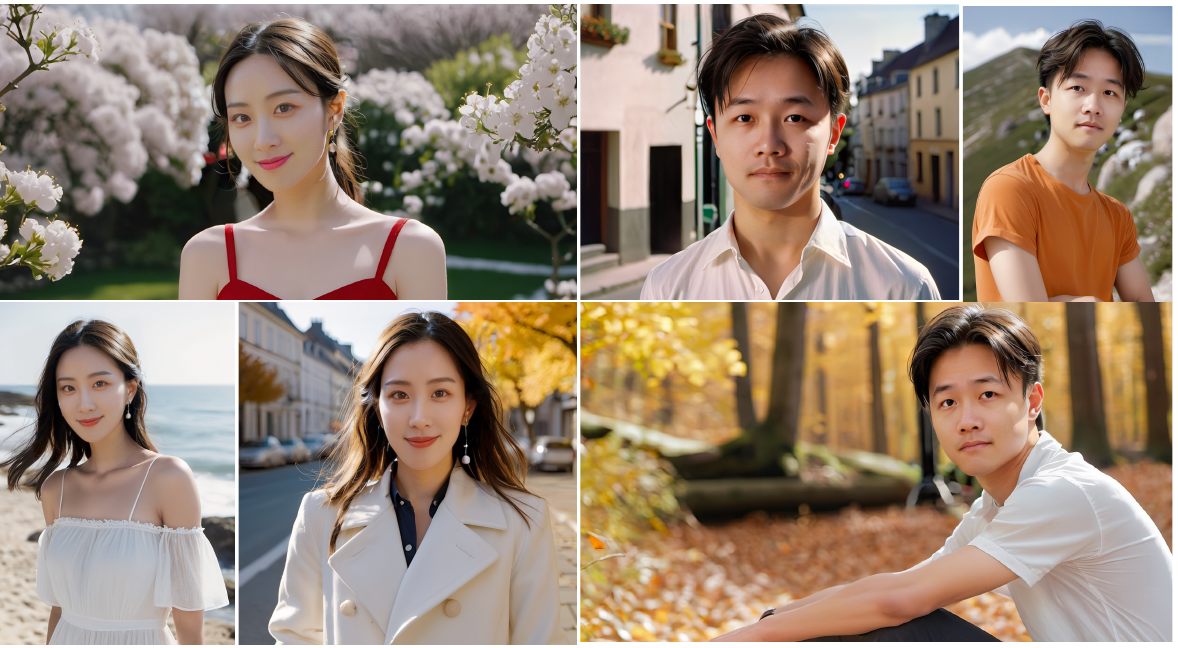}
	\caption{The generation results of EasyPhoto. (The template images are generated by SDXL model.)}\label{fig:sdxl}
\end{figure}

In this section, we present some generation results of EasyPhoto as shown in Fig. \ref{fig:res1} - Fig. \ref{fig:sdxl}. EasyPhoto now empowers users to generate AI portraits with various styles, multiple user IDs using provided templates, or even generate templates using the SD model (SDXL is also supported). These results demonstrate the capabilities of EasyPhoto in producing high-quality and diverse AI photos. To try out EasyPhoto yourself, you can install it as an extension in SD-WebUI or use the \href{https://help.aliyun.com/document_detail/2567864.html}{PAI-DSW} to start EasyPhoto within just 3 minutes. For more detailed information, you can refer to our repository.

\section{Conclusion}

In this paper, we present EasyPhoto, a WebUI plugin designed for the generation of AI photos. Our approach utilizes the base stable diffusion model alongwith a user-trained LoRA model, which is capable of producing high-quality and visually resemble photo results. The model includes face preprocessing techniques and two inpaint-based diffusion processes to address the issues of identity loss and boundary artifacts. Notably, EasyPhoto now allows users to generate AI photos with multiple users and offers the flexibility to choose from various styles. SDXL model is also adopted to generate more realistic and diverse infer template. In the future, we will extend the proposed algorithm by incorporating the concept of an "anything id." This means that the preprocess process of face can also be applied to any object region using powerful general models. We are actively working on making the code publicly available, allowing for seamless integration into EasyPhoto.

\section*{Acknowledgments}
We thank all the authors of the algorithms used in EasyPhoto for their contributions in the github community. We also thank the Modelscope Team that integrates the SOTA models and provides toolkit for quick model use. Many thanks to our family members and friends to share their photos.

\bibliography{easyphoto}
\bibliographystyle{acl_natbib}

\end{document}